\documentclass[letterpaper, 10 pt, conference]{ieeeconf}  

\IEEEoverridecommandlockouts   
\usepackage{tikz}
\usetikzlibrary{arrows.meta, calc}

\usepackage[T1]{fontenc}
\usepackage[utf8]{inputenc}
\usepackage{cite}
\usepackage{amsmath,amssymb,amsfonts}

\usepackage{amsthm}
\usepackage{graphicx,subcaption}
\usepackage{booktabs}
\usepackage{xcolor}
\usepackage{xspace}
\usepackage{url}
\usepackage{siunitx}
\usepackage[hidelinks]{hyperref}   

\graphicspath{{figures/}}

\newcommand{\ygg}{\textsc{Yggdrasil}\xspace}

\newcommand{\vanilla}{\emph{vanilla}\xspace}
\newcommand{\chocolate}{\emph{chocolate}\xspace}
\newcommand{\saffron}{\emph{saffron}\xspace}
\newcommand{\mint}{\emph{mint}\xspace}
\newcommand{\dsg}{\emph{dsg}\xspace}
\newcommand{\hide}[1]{}

\theoremstyle{definition}

\begin{document}

\title{\LARGE \bf
Yggdrasil: a Layer-First 3D Scene Graph for Real-Time Querying
}

\author{Arshia Akhavan$^{1}$, Ermanno Bartoli$^{2}$, Afnan Algharbi$^{1}$,\\
Alireza Hoseinpur$^{3}$, Iolanda Leite$^{2}$, and Bryan Donyanavard$^{1}$%
\thanks{$^{1}$Department of Computer Science, San Diego State University
        {\tt\small\{aakhavan3824, aalgharbi5447, bdonyanavard\}@sdsu.edu}}%
\thanks{$^{2}$Division of Robotics,
        Perception and Learning, KTH Royal Institute of Technology {\tt\small\{bartoli, iolanda\}@kth.se}}
\thanks{$^{3}$Department of Computer Science,
        University of Illinois Chicago
        {\tt\small ahose@uic.edu}}%
}

\maketitle
\thispagestyle{empty}
\pagestyle{empty}
\bstctlcite{IEEEexample:BSTcontrol}

\begin{abstract}
Robotic agents use 3D scene graphs (3DSG) to perform tasks ranging from scene
understanding to scene interaction. Although an extensive body of work addresses
scene graph generation, little attention has been paid to optimizing the graph for
consumption, which leaves state-of-the-art perception pipelines to work around their
own scene graph and to pay a latency cost that does not fit the real-time budget a
perception loop runs on.

We present \ygg, the first 3D scene graph designed to be
efficient for both generation and consumption: a layer-first hierarchical graph built
from generic nodes, edges, and layers, which expresses the representations existing
pipelines already produce, indoor or outdoor, flat or hierarchical, while natively
answering the positional and semantic queries downstream tasks issue.

Against a published DSG baseline, \ygg answers queries up to $121\times$ faster, and
every query we measure falls between 2 and 127\,\si{\micro\second}, three to five
orders of magnitude inside the 200\,ms keyframe budget a 3DSG consumer lives in, on both a
workstation and embedded class device. We integrate \ygg into three
published pipelines spanning human trajectory prediction, object-goal navigation, and
human-aware motion planning, where it removes up to 99\% of the time each spends on
its scene graph. The implementation, benchmark harness, and all three integrations
are available online~\cite{yggdrasil_artifact}.
\end{abstract}

\vspace{0.67ex}

\section{Introduction}
\label{sec:intro}

Robots operating in real environments must build a model of their
surroundings and interrogate it fast enough to act on what they find. That model
draws on the geometric maps produced by SLAM pipelines~\cite{cadena2016past} and the
semantic content associated with them~\cite{garg2020semantics}, and the
structure holding all of it together is commonly called a 3D scene graph
(3DSG)~\cite{armeni2019scenegraph}.

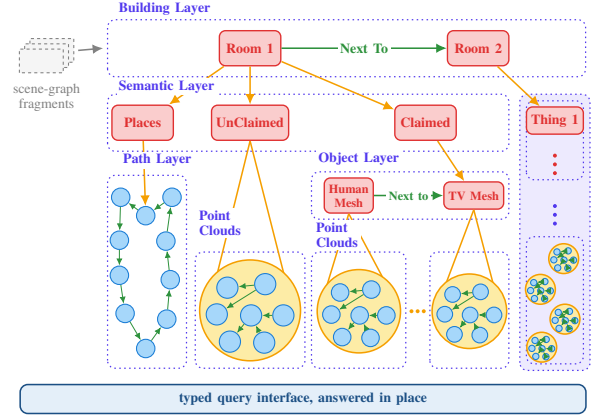
\begin{figure}[!t]
  \centering
  \resizebox{\columnwidth}{!}{
\definecolor{ygsemfill}{RGB}{250,205,205}
\definecolor{ygsemline}{RGB}{224,49,49}
\definecolor{ygmetfill}{RGB}{166,215,250}
\definecolor{ygmetline}{RGB}{28,126,214}
\definecolor{yglayer}{RGB}{81,56,238}
\definecolor{yghier}{RGB}{245,159,0}
\definecolor{ygrel}{RGB}{45,145,60}
\definecolor{ygcloudfill}{RGB}{253,236,166}
\definecolor{ygcolfill}{RGB}{237,234,254}
\definecolor{ygapi}{RGB}{27,79,132}
\definecolor{ygapifill}{RGB}{228,239,249}
\definecolor{yggrey}{RGB}{130,130,130}

\begin{tikzpicture}[
  font=\sffamily,
  x=1cm,y=1cm,
  sem/.style={rounded corners=4pt,draw=ygsemline,line width=0.9pt,
              fill=ygsemfill,text=ygsemline,font=\bfseries\small,
              inner sep=3pt,minimum width=1.4cm,minimum height=0.8cm,align=center},
  metric/.style={circle,draw=ygmetline,line width=0.7pt,fill=ygmetfill,
                 inner sep=0pt,minimum size=4.2mm},
  layerbox/.style={rounded corners=5pt,draw=yglayer,line width=0.8pt,
                   dash pattern=on 1pt off 2.2pt},
  layerlab/.style={text=yglayer,font=\bfseries\small,anchor=west,align=left,
                   fill=white,inner sep=1.5pt},
  hier/.style={draw=yghier,line width=1pt},
  hierarrow/.style={hier,-{Latex[length=2.4mm,width=2.2mm]}},
  rel/.style={draw=ygrel,line width=1pt,-{Latex[length=2.4mm,width=2.2mm]}},
  relthin/.style={draw=ygrel,line width=0.6pt,-{Latex[length=1.6mm,width=1.4mm]}},
  cloud/.style={draw=yghier,line width=1pt,fill=ygcloudfill},
  frag/.style={draw=yggrey,fill=black!4,line width=0.7pt,
               dash pattern=on 2pt off 2pt},
  api/.style={rounded corners=5pt,draw=ygapi,fill=ygapifill,text=ygapi,
              line width=1pt,font=\bfseries\small,align=center},
]

\newcommand{\ygcloud}[2]{%
  \begin{scope}[shift={#1},scale=#2,transform shape]
    \draw[cloud] (0,0) circle[radius=1];
    \node[metric] (c1) at (-0.45, 0.45) {};
    \node[metric] (c2) at ( 0.30, 0.52) {};
    \node[metric] (c3) at (-0.66,-0.08) {};
    \node[metric] (c4) at ( 0.02,-0.10) {};
    \node[metric] (c5) at ( 0.66,-0.10) {};
    \node[metric] (c6) at (-0.40,-0.62) {};
    \node[metric] (c7) at ( 0.28,-0.64) {};
    \draw[relthin] (c2) -- (c1);
    \draw[relthin] (c2) -- (c3);
    \draw[relthin] (c5) -- (c4);
    \draw[relthin] (c4) -- (c6);
    \draw[relthin] (c7) -- (c4);
  \end{scope}}

\draw[frag] (-1.62,7.28) rectangle (-0.62,7.82);
\draw[frag] (-1.74,7.16) rectangle (-0.74,7.70);
\draw[frag] (-1.86,7.04) rectangle (-0.86,7.58);
\node[text=yggrey,font=\small,anchor=north,align=center] at (-1.24,6.92)
  {scene-graph\\fragments};
\draw[draw=yggrey,line width=1pt,-{Latex[length=2.4mm,width=2.2mm]}]
  (-0.52,7.55) -- (0.05,7.60);

\draw[layerbox] (0.15,6.95) rectangle (11.05,8.25);
\node[sem] (room1) at (3.40,7.60) {Room 1};
\node[sem] (room2) at (8.60,7.60) {Room 2};
\draw[rel] (room1) -- node[midway,fill=white,inner sep=1.5pt,
            text=ygrel,font=\bfseries\small] {Next To} (room2);

\draw[layerbox] (0.15,5.25) rectangle (9.30,6.55);
\node[sem] (places)  at (0.95,5.90) {Places};
\node[sem] (unclaim) at (3.40,5.90) {UnClaimed};
\node[sem] (claim)   at (7.40,5.90) {Claimed};
\draw[hierarrow] (room1) -- (places);
\draw[hierarrow] (room1) -- (unclaim);
\draw[hierarrow] (room1) -- (claim);

\fill[ygcolfill,rounded corners=5pt] (9.55,0.25) rectangle (11.15,6.55);
\draw[layerbox] (9.55,0.25) rectangle (11.15,6.55);
\draw[layerbox] (9.70,4.60) rectangle (11.00,6.40);
\node[sem,minimum width=1.0cm,minimum height=0.62cm] (thing1) at (10.35,5.95) {Thing 1};
\foreach \y in {4.75,4.95,5.15}{\fill[ygsemline] (10.35,\y) circle[radius=0.045];}
\foreach \y in {3.60,3.80,4.00}{\fill[yglayer] (10.35,\y) circle[radius=0.045];}
\draw[layerbox] (9.70,0.40) rectangle (11.00,3.30);
\ygcloud{(10.55,2.80)}{0.34}
\ygcloud{(10.02,2.10)}{0.34}
\ygcloud{(10.58,1.42)}{0.34}
\ygcloud{(10.05,0.78)}{0.34}
\draw[hierarrow] (room2) -- (thing1);

\draw[layerbox] (0.15,0.30) rectangle (1.95,4.70);
\node[metric] (p1)  at (0.45,4.20) {};
\node[metric] (p2)  at (1.62,4.20) {};
\node[metric] (p3)  at (1.02,3.78) {};
\node[metric] (p4)  at (0.42,3.22) {};
\node[metric] (p5)  at (0.42,2.42) {};
\node[metric] (p6)  at (0.55,1.55) {};
\node[metric] (p7)  at (1.02,0.75) {};
\node[metric] (p8)  at (1.50,1.42) {};
\node[metric] (p9)  at (1.58,2.32) {};
\node[metric] (p10) at (1.48,3.20) {};
\foreach \a/\b in {p3/p1,p2/p3,p1/p4,p4/p5,p5/p6,p6/p7,p7/p8,p8/p9,p9/p10,p10/p2}
  {\draw[relthin] (\a) -- (\b);}
\draw[hierarrow] (places) -- (p3);

\draw[layerbox] (2.15,0.30) rectangle (4.65,2.95);
\draw[hier] (unclaim.south) -- (2.55,2.05);
\draw[hier] (unclaim.south) -- (4.28,2.02);
\ygcloud{(3.40,1.62)}{1.15}

\draw[layerbox] (4.80,3.70) rectangle (9.30,4.78);
\node[sem,minimum width=1.1cm,font=\bfseries\footnotesize] (human) at (5.65,4.24) {Human\\Mesh};
\node[sem,minimum width=1.1cm,minimum height=0.62cm,font=\bfseries\footnotesize] (tv) at (8.50,4.24) {TV Mesh};
\draw[rel] (human) -- node[midway,fill=white,inner sep=1.5pt,
            text=ygrel,font=\bfseries\footnotesize] {Next to} (tv);
\draw[hierarrow] (claim) -- (tv);

\draw[layerbox] (4.80,0.30) rectangle (6.95,2.95);
\draw[hier] (human.south) -- (5.10,2.02);
\draw[hier] (human.south) -- (6.62,2.00);
\ygcloud{(5.88,1.60)}{0.95}
\foreach \x in {7.08,7.22,7.36}{\fill[yghier] (\x,1.60) circle[radius=0.045];}
\draw[layerbox] (7.50,0.30) rectangle (9.30,2.95);
\draw[hier] (tv.south) -- (7.76,1.95);
\draw[hier] (tv.south) -- (9.02,1.93);
\ygcloud{(8.40,1.60)}{0.85}

\node[api,minimum width=12.9cm,minimum height=0.66cm,anchor=south west]
      at (-1.86,-0.75) {typed query interface, answered in place};

\node[layerlab] at (0.35,8.50) {Building Layer};
\node[layerlab] at (0.35,6.72) {Semantic Layer};
\node[layerlab] at (0.45,5.05) {Path Layer};
\node[layerlab] at (2.18,3.62) {Point\\Clouds};
\node[layerlab] at (4.90,5.08) {Object Layer};
\node[layerlab] at (4.85,3.35) {Point\\Clouds};
\end{tikzpicture}}
  \caption{\ygg's layer-first architecture: a DAG of layers, each an isolated graph,
  with nesting relations connecting nodes across layers that share an edge in the DAG.
  A typed query interface answers semantic and spatial queries against this store in
  place, with no intermediate conversion.}
  \label{fig:arch}
\end{figure}

A 3DSG sits between two very different halves of the pipeline: \emph{upstream}, a
construction stage that populates 3DSG from raw sensor data, and \emph{downstream},
applications that consume it to plan, navigate, or answer questions about the scene.
3DSG's purpose is to hand the downstream task \emph{actionable}
information~\cite{kim2019sparse3d, armeni2019scenegraph}, a contract expressed in
practice as the \emph{queries} that application runs against it, such as the nearest object, the observable sub-scene, the objects carrying a given
label, and the shortest path query.

A large body of work addresses the upstream
side~\cite{bae2023survey3dscenegraphs, wald2020learning3d, wu2021scenegraphfusion, hughes2022hydra}, designing representations that perception pipelines can populate
accurately and in real time. Far less attention has gone to the downstream, where
the graph would have to be designed around the queries and information retrieval.
\hide{
To the best of our knowledge, no practical scene graph technology optimizes for consumption.
}
Downstream pipelines therefore have to route \emph{around} their scene graph.
Some serialize it for a language model to delegate the query~\cite{rana2023sayplan}, which works
but is slow relative to the control loop~\cite{mahdi2026scout, zhou2025fsrvln} and
can face a context window bottleneck for large scenes~\cite{linok2025beyondbarequeries, werby2025keysg}.
Others keep the query
algorithmic but re-materialize the graph inside a general-purpose library such as
NetworkX~\cite{hagberg2008networkx} and query the copy~\cite{loo2024osg, gorlo2024trajectory}. That presumes an immutable graph: when the scene is updated
while navigation or planning runs, the conversion is re-paid on every frame, and the
copy discards the layering and grounding that distinguish a 3DSG from a generic graph.\hide{fixme{new line}}
Both are workarounds for the same missing piece, a scene graph that serves upstream
construction while natively answering what downstream robotics asks of it. Recent
work has begun to name the gap~\cite{chang2023dlite} and to propose exposing 3DSGs
through a structured query interface for tool-augmented
reasoning~\cite{ra2025structuredinterface}; we pursue that goal one level lower, at
the storage substrate, so that constructors and consumers share a single queryable
graph instead of each building its own.

We present \ygg, the first 3D scene graph designed to be efficient for both
generation and consumption (Fig.~\ref{fig:arch}). Built from generic nodes, edges,
and layers, it expresses the representations existing pipelines already produce,
indoor or outdoor, flat or hierarchical, while natively answering the semantic and
spatial queries downstream applications issue, at practical latency and with no
intermediate conversion step. This paper contributes:

\begin{enumerate}
  \item \textbf{A scene graph designed for consumption}, in which the layer
        hierarchy is a directed acyclic graph rather than a stack, paired with a
        query API spanning pattern matching, nearest neighbour, field of view, and
        traversal, backed by per-layer and per-node spatial indexes that \ygg
        maintains automatically (Sections~\ref{sec:design}, \ref{sec:queries}).
  \hide{\item \textbf{An ablation isolating where the speed comes from}, on the
        \texttt{uHumans2} office graph~\cite{rosinol2021kimera}
        against a DSG~\cite{sparkdsg} baseline and four \ygg
        configurations, on a workstation and on a Jetson Orin Nano
        (Section~\ref{sec:eval}).}
  \item \textbf{Three integrations spanning the perception pipeline}, in published
        tools for human trajectory prediction, object-goal navigation, and
        human-aware motion planning, covering diverse use cases such as: downstream-only use, construction and
        consumption together inside a control loop, and augmentation of an existing
        graph (Section~\ref{sec:integration}).
\end{enumerate}

\section{Related Work and Background}
\label{sec:related}

\subsection{3D Scene Graphs: Definitions, Hierarchy, and Layering}

Although no single definition of a 3DSG is universally agreed
upon~\cite{bae2023survey3dscenegraphs, rotondi2026survey}, the term
consistently denotes the storage backbone of a perception pipeline. A recent
survey~\cite{rotondi2026survey} unifies the competing formulations as sets of nodes
and edges plus a \emph{grounding} that anchors each node to the
metric map, \emph{feature maps} that attach arbitrary attributes, and an
\emph{optional} labeling function sorting nodes into hierarchical \emph{layers}. A
graph lacking the labeling function is considered flat.

In a 3DSG, nodes are spatial concepts, most often objects, and edges are the relationships between
them~\cite{rotondi2026survey}. Nodes carry geometric attributes (centroid,
bounding box, shape) and semantic ones (labels, materials, embeddings). Flat designs
such as the 3D semantic scene graph (3DSSG)~\cite{wald2020learning3d} and
ConceptGraphs~\cite{gu2024conceptgraphs} stop
there and invest in richer attributes, whereas hierarchical ones add layers of
increasing abstraction, commonly geometry $\rightarrow$ objects $\rightarrow$ places
$\rightarrow$ rooms $\rightarrow$ building~\cite{armeni2019scenegraph, rosinol2020dynamic, bavle2022situational, werby2024hierarchical}. Hierarchy both
mirrors the human mental model of an environment~\cite{rosinol2021kimera, bavle2025sgraphs2} and aids inference by narrowing the search span along
ancestry~\cite{hughes2024foundations}, since an object in one room need not be sought
among those of another. Within this family, the DSG line~\cite{rosinol2020dynamic, rosinol2021kimera, hughes2022hydra, hughes2024foundations} adds dynamic entities such
as humans, S\nobreakdash-Graphs~\cite{bavle2022situational, bavle2023sgraphs, bavle2025sgraphs2} couples the hierarchy to SLAM optimization, and
OpenGraph~\cite{deng2024opengraph} adapts the layering to large-scale \emph{outdoor}
environments.

\subsection{Scene Graph Construction and Consumption}

Work on 3DSGs falls into two threads. \textbf{Construction (upstream)}, called
\emph{generation} in robotics and \emph{learning} in computer
vision~\cite{rotondi2026survey}, builds the graph from sensor data, and spans learned
semantic prediction from point clouds and RGB-D~\cite{wald2020learning3d, wu2021scenegraphfusion}, open-vocabulary variants~\cite{koch2024open3dsg, gu2024conceptgraphs}, and real-time, SLAM-coupled construction of layered
graphs~\cite{hughes2022hydra, deng2024opengraph, bartoli2026gesto}; its output is the
graph itself. \textbf{Consumption (downstream)} instead uses an existing graph to
solve a task: embodied question answering (GraphEQA~\cite{saxena2024grapheqa}),
LLM-based task and motion planning (SayPlan~\cite{rana2023sayplan},
Taskography~\cite{agia2022taskography}, DELTA~\cite{liu2025delta}), object-goal
navigation (SG-Nav~\cite{yin2024sg}, HOV-SG~\cite{werby2024hierarchical},
OSG~\cite{loo2024osg}), and predictive or
interactive tasks such as human-trajectory
prediction~\cite{gorlo2024trajectory} and interactive object
search~\cite{honerkamp2024language}. Neither thread designs the store around the
queries run against it. \ygg does, and answers them in place rather than from a
converted copy.

\section{Core Design}
\label{sec:design}

\ygg is a layer-first hierarchical 3D scene graph built from generic nodes, edges,
and layers. Because none of the three is bound to a predefined vocabulary, a single
\ygg graph can express the representations that existing pipelines already produce,
indoor or outdoor, flat or hierarchical, while still scoping a query to a single
room.

Architecturally, \ygg is a cluster of graphs. Each graph is a \emph{layer}, and the
layers together form a directed acyclic graph (DAG). Layers are almost pairwise
isolated, in that no edge joins nodes from different layers; the single exception is
the \emph{nesting} relationship described below. Each layer represents one level of
abstraction over the entire scene and holds one kind of information: a physical
layer may carry positional data, a semantic layer the attributes of each object, and
a human-interaction layer only the interactions taking place between the people in
the scene.

Nodes represent different aspects of an object, ranging from physical
representations such as a point in a point cloud or a surface in a mesh, to semantic
ones such as its type. Edges represent a relationship between two abstractions: an
edge between the physical nodes of a human and a mug might encode \emph{closeness},
while an edge between their semantic nodes might encode an action such as
\emph{holding}.

Where two layers share a directed edge in the DAG, the source layer
\emph{encompasses} the destination and sits at a higher level of abstraction. This
licenses the one cross-layer relation \ygg admits: node \texttt{n1} in layer
\texttt{l1} may \emph{nest} node \texttt{n2} in layer \texttt{l2} exactly when
\texttt{l1} encompasses \texttt{l2}. Some form of nesting appears in every 3D scene
graph that exposes hierarchy through layers~\cite{rosinol2020dynamic}. What
distinguishes \ygg is that prior work treats the layer hierarchy as a \emph{stack},
with \texttt{l1} the parent of \texttt{l2}, \texttt{l2} the parent of \texttt{l3},
and so on, whereas \ygg's DAG lets a layer parent more than one child.

This matters most where a stack cannot express the scene at all. In outdoor graphs
two layers often cannot be ranked, because neither stands above the other.
OpenGraph~\cite{deng2024opengraph} keeps one layer for the road lane graph and
another for object instances; under a stack schema one of them must be placed above
the other even though no parent-child relationship holds between a lane and an
object. \ygg's DAG materializes the true relationship by letting both descend from a
common parent, which in OpenGraph's case is the \emph{segment} layer that partitions
the environment by lane connectivity.

The same freedom pays off indoors in two further ways, both of which
Section~\ref{sec:eval} quantifies. Objects that are categorically the same but
semantically distinct can be separated: the same level of abstraction for objects in
different rooms can live in one layer per room, so that an object in one room need
never be compared against nodes in another and a query's search span is bounded by
the room rather than by the building. The finer granularity also means more layers
holding fewer nodes each, which shortens query time~\cite{hughes2024foundations} and
admits layer-specific optimizations that a single, heavily populated layer cannot
support: because \ygg builds indexes per layer and per node rather than globally, a
frequently queried room can be indexed on its own.

A single scene captures all three. Consider an office with one inventory room, where
positional queries and object detection matter, and meeting rooms where they do not.
A conventional hierarchical 3DSG models the whole office's physical extent in one
layer; \ygg can give each room's point cloud a layer of its own, so object detection
in the inventory is constrained to the inventory's objects rather than the entire
building. And knowing that the inventory is queried often, one can pay a small memory
price there for faster queries, which was previously unavailable because the cost of
such an optimization scaled with the physical layer of every object in the
building.

\begin{figure*}[!t]
  \centering
  \begin{subfigure}[b]{0.285\textwidth}
    \centering
    \includegraphics[width=\linewidth]{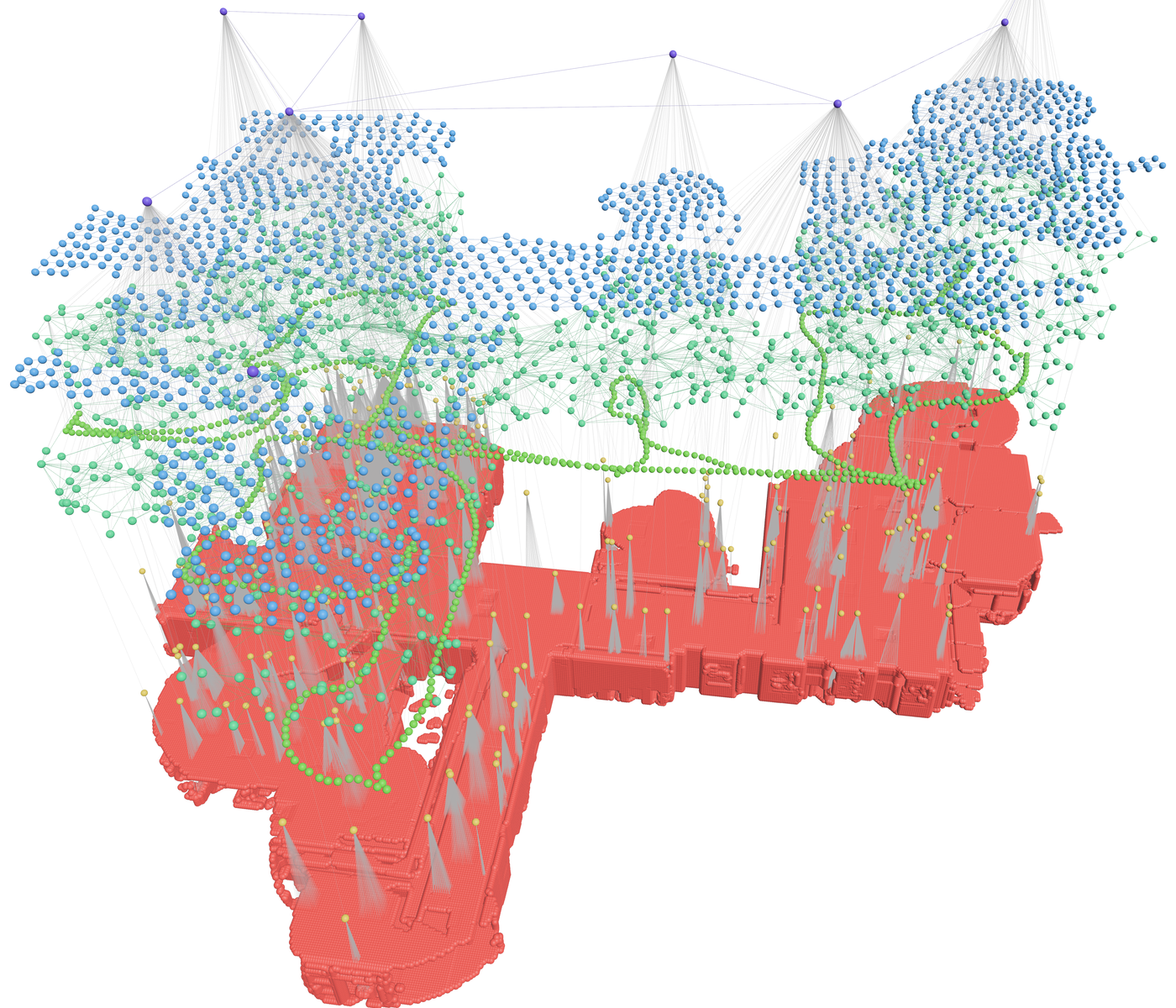}
    \caption{Example 3DSG in the DSG format.}
    \label{fig:ygg-dsg}
  \end{subfigure}
  \hfill
  \begin{subfigure}[b]{0.305\textwidth}
    \centering
    \includegraphics[width=\linewidth]{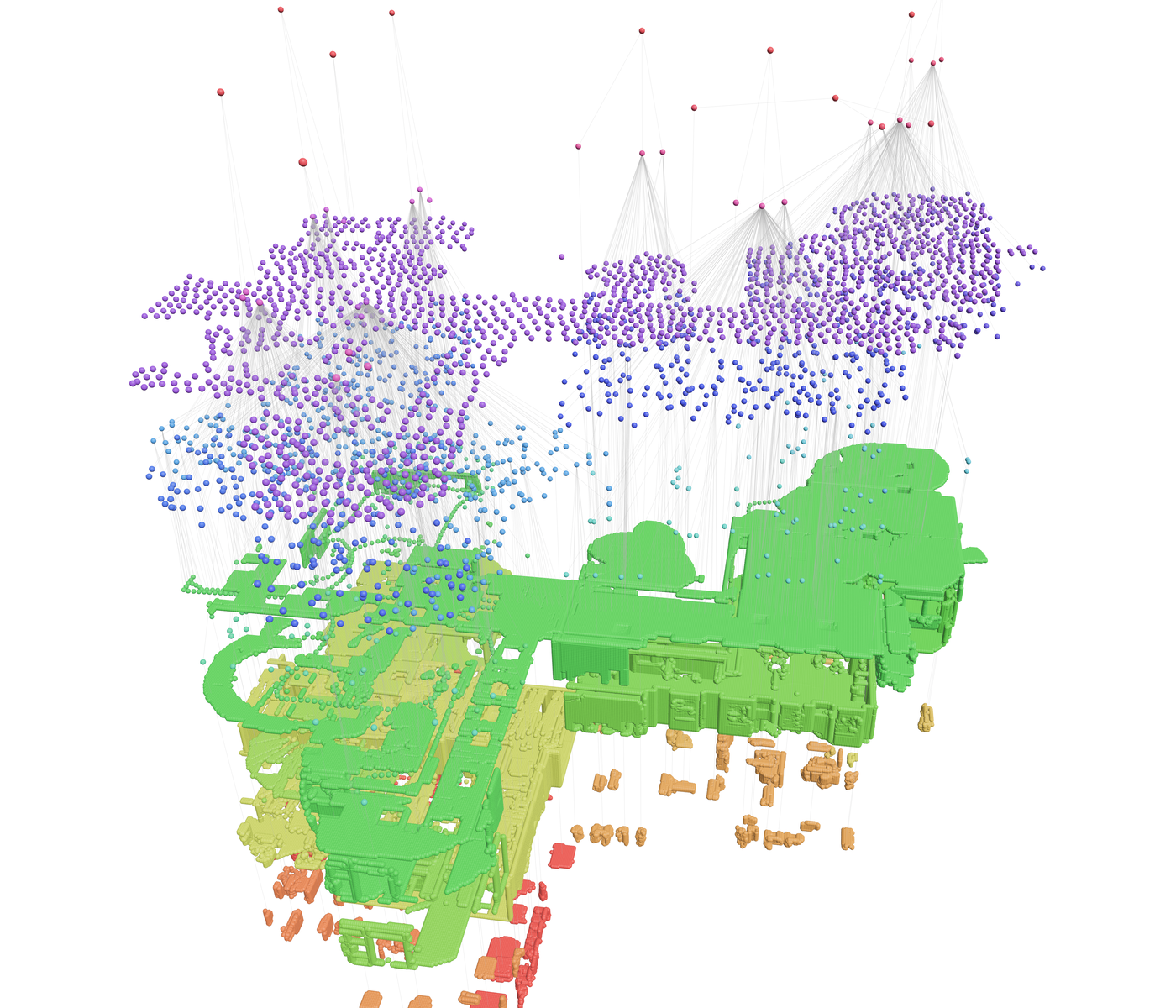}
    \caption{The same scene graph in \ygg's layer-oriented hierarchy.}
    \label{fig:ygg-layers}
  \end{subfigure}
  \hfill
  \begin{subfigure}[b]{0.375\textwidth}
    \centering
    \resizebox{\linewidth}{!}{
\definecolor{cOrig}{RGB}{202,80,42}
\definecolor{cOrigFill}{RGB}{251,231,222}
\definecolor{cYgg}{RGB}{27,79,132}
\definecolor{cYggFill}{RGB}{228,239,249}
\definecolor{cGhost}{RGB}{155,155,155}
\begin{tikzpicture}[
  x=1cm,y=1cm,font=\footnotesize,
  stage/.style={draw,line width=0.7pt,align=center,font=\scriptsize,
                inner sep=2pt,minimum height=0.62cm,minimum width=1.45cm},
  ostage/.style={stage,draw=cOrig,fill=cOrigFill,text=cOrig},
  ystage/.style={stage,draw=cYgg,fill=cYggFill,text=cYgg},
  gstage/.style={stage,draw=cGhost,fill=black!3,text=cGhost,
                 dash pattern=on 2pt off 2pt},
  ofl/.style={draw=cOrig,line width=0.8pt,-{Latex[length=1.8mm,width=1.6mm]}},
  yfl/.style={draw=cYgg,line width=0.8pt,-{Latex[length=1.8mm,width=1.6mm]}},
  note/.style={font=\scriptsize,align=center},
  api/.style={rounded corners=3pt,draw=cYgg,fill=cYggFill,text=cYgg,
              line width=0.8pt,font=\scriptsize,align=center},
]
\node[ostage] (sgA) at (0.85,3.92) {Spark-DSG};
\node[ostage,minimum width=1.85cm] (mrA) at (2.95,3.92) {NetworkX copy\\$+$ KD-tree};
\node[ostage] (qA)  at (5.00,3.92) {LP2\\queries};
\draw[ofl] (sgA) -- (mrA);
\draw[ofl] (mrA) -- (qA);
\node[note,text=cOrig,anchor=north] at (2.95,3.44)
  {rebuilt every iteration:\\\textbf{2\,681 ms} build $+$ \textbf{7\,747 ms} queries};
\node[ystage] (sgB) at (0.85,1.92) {Yggdrasil};
\node[gstage,minimum width=1.85cm] (mrB) at (2.95,1.92) {NetworkX copy\\$+$ KD-tree};
\node[ystage] (qB)  at (5.00,1.92) {LP2\\queries};
\begin{scope}[cOrig,line width=0.9pt,opacity=0.55]
  \draw (mrB.south west) -- (mrB.north east);
  \draw (mrB.north west) -- (mrB.south east);
\end{scope}
\draw[yfl,rounded corners=4pt]
  (sgB.east) -- (1.75,1.92) -- (1.75,2.48) -- (5.00,2.48) -- (qB.north);
\node[note,text=cYgg,anchor=north] at (2.95,1.44)
  {answered natively:\\\textbf{35 ms} build $+$ \textbf{31 ms} queries};
\node[api,minimum width=5.0cm,minimum height=0.48cm] at (2.95,0.50)
  {\textbf{10.4 s} $\rightarrow$ \textbf{66 ms} per trajectory (\textbf{158}$\times$)};
\end{tikzpicture}}
    \caption{The LP2 case study: \ygg removes the NetworkX and KD-tree mirror that the
    baseline rebuilds every iteration.}
    \label{fig:ygg-lp2}
  \end{subfigure}
  \caption{Porting a scene graph into \ygg. (a) and (b) show the same office scene in
  the DSG format and in \ygg's layer-oriented hierarchy (each color presents a layer); (c) shows what the change
  buys downstream in LP2.}
  \label{fig:exp}
\end{figure*}
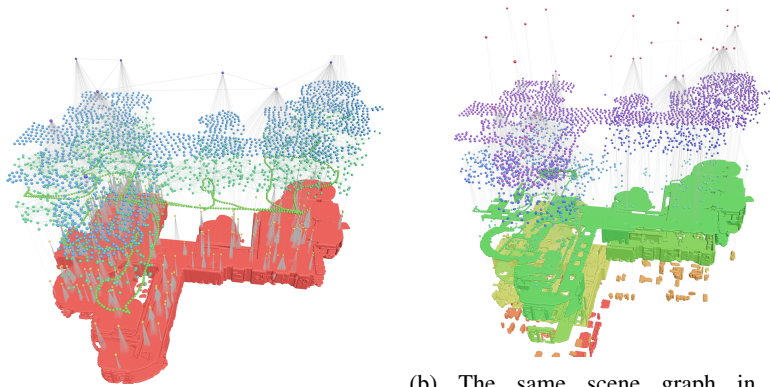

Two derived notions recur throughout the paper. The \emph{sub-scene graph} over a
node set \texttt{N} of a scene graph \texttt{sg} is the graph containing every node
in \texttt{N}, all edges and nesting relations among them, and all layers of
\texttt{sg}, retained even when empty, since preserving the layers matters for
operations such as merging and conflict resolution. The \emph{sub-tree} of a node is
then the sub-scene graph over the nodes nested under it, directly or through a chain
of nesting, and usually captures an entire object across levels of abstraction: in
Fig.~\ref{fig:arch}, the sub-tree of \texttt{Room 1} holds the room and every object
in it.

Nodes are generic in that, unlike scene graph technologies offering a fixed set of
node types~\cite{rosinol2020dynamic, loo2024osg, gorlo2024trajectory}, they can
encode any kind of information: each carries a list of features and an optional field
for \textbf{geometric coordinates}. A node may also track its \emph{bounding box},
the smallest axis-aligned cuboid containing every coordinate in its sub-tree, which
is useful for quick spatial estimates and is relied upon by several downstream
applications~\cite{gu2024conceptgraphs, maggio2024clio}. Bounding boxes are disabled
by default, because \ygg keeps every enabled box current as the graph changes, so
enabling one costs a small penalty on each update within that node's sub-tree. Edges
are generic in the same sense, bound to no predefined vocabulary, directed, and
carrying the source and destination of the relation they encode.

\section{Query APIs}
\label{sec:queries}

\ygg comes with a rich set of semantic and spatial queries suited to the variety
of scenarios downstream applications present.

\subsection{Pattern Matching}

Each node carries a set of features, essentially key-value pairs, while each edge
carries a source, a destination, and a description of the relationship between the
two. \ygg exposes \texttt{nodes\_matching} and \texttt{nodes\_having} to retrieve
nodes by a specific list of features or by feature keys alone, and
\texttt{edges\_from}, \texttt{edges\_to}, and \texttt{edges\_matching} to filter
edges on any of those three attributes. Every one of these queries can be run
either over the entire scene graph or over a single layer.

\subsection{Spatial and Traversal Queries}

The nearest-neighbor family, central to navigation and path planning, comes in
three flavors: \texttt{nearest\_nodes} returns at most \texttt{m} nodes closest to
a given coordinate \texttt{c}, \texttt{nodes\_within\_radius} restricts those to
the ones lying within a radius \texttt{r} of \texttt{c}, and
\texttt{nearest\_in\_subtree} restricts them instead to the sub-tree of a root node
\texttt{r}. The first two run over the whole scene graph or over a single layer,
while the third is an inter-layer query by definition and so runs only over the
graph.

Because these queries are issued frequently, \ygg can build indexes for them,
trading memory for runtime. Layer-level indexing tracks every coordinate stored in
a frequently queried layer and serves the first two flavors, while node-level
indexing tracks the coordinates in a node's sub-tree and serves the third. \ygg
maintains these indexes itself and updates them whenever the graph changes in a way
that concerns them, which is also why enabling one imposes a small runtime penalty
on updates. Both are backed by a
\texttt{kd-tree}~\cite{bentley1975kdtree, friedman1977kdtree}.

The field-of-view query serves downstream tasks that perform traversal or object
detection. Given an observer, \ygg runs a fast frustum check over the nodes lying
in the sub-tree of a root node \texttt{r}, which lets a caller confine the result
to the room an agent currently occupies rather than the whole scene. Where
node-level indexing is available, the executor opts into the index-based
implementation.

Beyond these scene-graph--specific queries, \ygg supports ordinary graph traversal
and allows any standard algorithm, such as shortest path or breadth- and
depth-first search, to be run over the graph. It also exposes a \texttt{subgraph}
query, which returns the subgraph rooted at a given node \texttt{r}.

\section{Evaluation}
\label{sec:eval}

We evaluate \ygg by porting a published DSG scene graph into it and measuring the
cost of the queries a downstream application issues against it. The evaluation is an
\emph{ablation ladder}: each target adds one \ygg feature to the one before it, so
that implementation, layer-first design, and live spatial indexes can be credited
separately, along with what each of them costs in memory.

\subsection{Dataset}
\label{sec:eval:dataset}

All measurements use the \textbf{uHumans2 office} scene graph distributed with
Spark\nobreakdash-DSG~\cite{sparkdsg}: a single-story office built by
Hydra~\cite{hughes2022hydra, hughes2024foundations} and semantically segmented
against the ADE20K label set~\cite{zhou2017ade20k}. It holds 3,532 nodes and 11,012
edges across four DSG layers, one building, nine rooms, 1,130 places and 1,472
mesh-places, 705 agent poses, and 215 objects, beside a side array of 354,885 mesh
vertices into which each object indexes to claim its own geometry.

Two properties of the data shape the experiments that follow. First, most of the mesh
is unclaimed, with 313,522 vertices (88.3\%) belonging to no object and a further
7.89\% claimed by more than one, so recovering an object from its geometry needs a
tie-breaking rule, for which we take the smallest object id to keep the output
reproducible. Second, the stored bounding boxes are unreliable as geometric bounds:
126 of the 215 objects have claimed vertices lying \emph{outside} their own box. That
divergence is most likely a consequence of maintaining the boxes separately from the
graph itself, which \ygg avoids by keeping every enabled box truthful on each
update.

\subsection{Targets}
\label{sec:eval:targets}

The evaluation uses five targets: the published baseline, and four ports into \ygg
that each differ from their predecessor in a single respect.

\begin{enumerate}
  \item \textbf{\dsg}: the baseline scene graph as provided by the office dataset.
        Its scene consists of four semantic layers, holding objects and agents,
        places and structures, rooms, and the building, together with a mesh array
        that holds the mesh representation of the entire scene \emph{outside} the
        scene graph.

  \item \textbf{\vanilla}: the naive port of DSG into \ygg, with no change to the
        internal design of the scene graph, which isolates \ygg's non-architectural
        differences such as implementation language (DSG's C++ against \ygg's Rust)
        and other runtime factors. Since \ygg is a single scene graph, DSG's separate
        mesh array could not be ported; we instead gave each object a node in a mesh
        layer holding its mesh group.

  \item \textbf{\chocolate}: the \emph{smart} port, in which we redesign the scene
        graph into a \ygg-friendlier one (Fig.~\ref{fig:exp}), and which therefore isolates the benefit
        of \ygg's new concepts, namely the DAG of layers and the finer node
        hierarchy. Because practical queries are typically scoped to a single room,
        and because the unclaimed mesh of Section~\ref{sec:eval:dataset} adds noise to
        queries that should retrieve objects, we allocate four layers per room:
        \emph{room},
        \emph{claimed-object}, \emph{claimed-mesh}, and \emph{unclaimed-mesh}. Each
        \emph{room} layer holds two nodes, \emph{claimed} and \emph{unclaimed}, both
        nested under that room's node in the building layer. Unclaimed meshes go
        into the \emph{unclaimed-mesh} layer, nested under the \emph{unclaimed} node
        of their room, while objects go into \emph{claimed-object} under the
        \emph{claimed} node, each nesting its own mesh node in \emph{claimed-mesh}.
        The building layer is unchanged from \vanilla and \dsg.

  \item \textbf{\mint}: structurally identical to \chocolate, differing only in that
        indexes are enabled, node indexing for each object's semantic node and layer
        indexing for each room's \emph{claimed-mesh} layer. This target isolates the
        benefit of \ygg's live spatial indexes.

  \item \textbf{\saffron}: inherits \mint and differs only in the granularity of the
        physical layer, giving every point of the mesh its own node rather than one
        node holding an object's whole mesh array. This isolates the cost of a
        fine-grained physical representation.
\end{enumerate}

\subsection{Query Workload}
\label{sec:eval:workload}

We run three families of queries (Q1-3), with several flavors within each, to expose the
trade-offs that come with each implementation. Every spatial query is issued from 64
reference points per room, drawn from a fixed seed, for each of the nine rooms.
Throughout, \dsg and \vanilla use the brute-force implementation, scanning every
candidate in scope, while \chocolate, \mint, and \saffron call the corresponding
\ygg API scoped to the room the reference point falls in, which the index serves on
\mint and \saffron. On all 576 spatial queries per configuration, all \ygg's flavors
return exactly the answer the \dsg returns, so the latencies below compare
identical results.

\textbf{Q1a} asks which object lies at minimum Euclidean distance from a given 3D
point, using only the geometry stored in the graph and breaking ties toward the
smaller node id. \textbf{Q1b} retrieves every object with a point within a given
distance of a reference point, at radii of 0.25, 0.5, 1, 2, and 4\,m, the largest of
which covers almost all of the smaller rooms. \textbf{Q2} asks what subset of the
graph is visible to an observer, each of the 64 per room given a half-angle field of
view between $30^\circ$ and $70^\circ$ and a range between 0.3 and 8\,m.
\textbf{Q3} returns every object carrying a given label, where \dsg allows one label
per node and \ygg a set of them; the dataset holds 27 labels with frequencies from 1
to 40, and we query one from each of four classes: \emph{frequent} (\emph{light},
40), \emph{mid} (\emph{shelf}, the median at 5), \emph{rare} (\emph{bulletin},
once), and \emph{absent}, a label occurring nowhere.

\subsection{Experimental Setup}
\label{sec:eval:setup}

Experiments were run on two platforms: a 32-core AMD Ryzen Threadripper PRO 3975WX
workstation with 251\,GiB of memory, and an NVIDIA Jetson Orin Nano Super developer
kit with six Arm Cortex-A78AE cores and 7.4\,GiB of unified memory in the uncapped
\texttt{MAXN\_SUPER} power mode. Each query is run five times on each machine for
each target and we report the average.

\subsection{Results}
\label{sec:eval:results}

\begin{figure*}[!t]
  \centering
  \includegraphics[width=\textwidth]{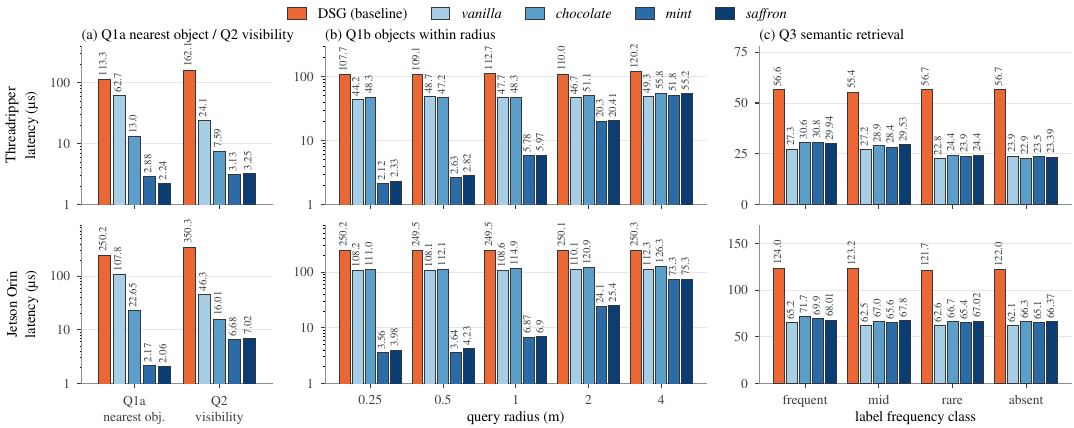}
  \caption{Query latency in \si{\micro\second} on the uHumans2 office scene graph,
  averaged over five runs, every value printed on its bar. Columns are query families;
  rows are the two machines of Section~\ref{sec:eval:setup}, each with its own
  vertical scale. \textbf{Note the logarithmic axis in (a) and (b).} Bars follow the
  legend order.}
  \label{fig:eval}
\end{figure*}

\begin{table}[!t]
\caption{Static cost of each target on the uHumans2 office graph.}
\label{tab:static}
\centering
\small
\setlength{\tabcolsep}{6pt}
\begin{tabular}{l r r r}
\toprule
target & nodes & memory (MB) & update latency (\si{\micro\second}) \\
\midrule
\dsg       & 3,532$^{\dagger}$ & 40.3           & 7.54\\
\vanilla   & 3,746             & 25.6           & 7.01 \\
\chocolate & 3,778             & 33.7           & 6.91\\
\mint      & 3,778             & 70.9           & 12.7\\
\saffron   & 361,980           & 144.6          & 129 \\
\bottomrule
\end{tabular}

\vspace{2pt}
{\footnotesize $^{\dagger}$ plus the 354,885-vertex mesh array, which \dsg stores
outside the graph.}
\end{table}

Fig.~\ref{fig:eval} reports per-query latency for all five targets and
Table~\ref{tab:static} the static cost of each. The two rows have essentially the
same shape, indicating that \ygg's improvements are a property of the design rather
than of the host, and that they survive on a platform as constrained as the Jetson.

On Q1a and Q2, every rung of the ladder pays. Porting DSG unchanged into \ygg is
already 1.8--2.3$\times$ faster on nearest-object and 6.7--7.6$\times$ on
visibility, before any design change, which credits \ygg's data structures and
packed memory layout. Redesigning the graph adds a further 4.8$\times$ on Q1a and
about 3$\times$ on Q2, purely from reduced search span, since \chocolate gives every
room its own isolated stack of layers. The indexes buy another 5--10$\times$ on Q1a
and 2.4$\times$ on Q2 by running the search against the kd-tree rather than over raw
points. End to end, the fully featured targets answer Q1a in about
2\,\si{\micro\second} on both machines, 50$\times$ faster than \dsg on the
Threadripper and 121$\times$ on the Jetson, and cut Q2 by 52$\times$.

Q1b is the one query on which the layer redesign buys nothing: it must touch every
point in scope regardless, so room isolation prunes nothing while the extra layers
are still walked, leaving \chocolate marginally slower than \vanilla at the larger
radii. Here the indexes are the whole story, and their benefit falls with the
radius, from 23--31$\times$ over \chocolate at 0.25\,m to 1.1--1.7$\times$ at 4\,m,
the expected behaviour once the query ball covers the whole room.

Q3 flattens the ladder altogether. All four \ygg targets are indistinguishable and
constant across the frequency classes, each about twice as fast as \dsg, since label
matching is invariant to the design changes that separate them. \saffron is the
interesting case, holding 96$\times$ more nodes (Table~\ref{tab:static}) yet staying
in the same range, because the geometric nodes it adds carry no labels by
construction, which lets \ygg skip those layers altogether.

\subsection{Discussion}
\label{sec:eval:discussion}

Table~\ref{tab:static} shows what each rung costs to hold. \vanilla is not only
faster than \dsg but also 36\% smaller, and \chocolate's per-room layer stacks still
fit below the \dsg baseline at 33.7\,MB. The two features that do cost memory are the
live indexes, which add 37.2\,MB to \mint over \chocolate, and \saffron's point-level
physical layer, which brings the total to 144.6\,MB, 3.6$\times$ the \dsg baseline.

Aside from \vanilla, an ablation baseline, the three remaining \ygg implementations
answer queries well inside the real-time bounds of state-of-the-art perception
pipelines. Systems that run end to end, from raw sensor data through to a downstream
task, report explicit cycle budgets. Hydra, which builds a 3DSG online from RGB-D
data, targets keyframe rate (5\,Hz), which as its authors state ``implies a limit of
200\,ms for any of these processes''~\cite{hughes2024foundations}; on an Nvidia Xavier
NX, and on the same uHumans2 office scene we evaluate on, it reports $75 \pm 35$\,ms
for objects, $33 \pm 6$\,ms for places, and $55 \pm 41$\,ms for
rooms~\cite{hughes2022hydra}. Kimera reports per-frame meshes in under 20\,ms and a
global metric-semantic mesh at roughly 0.1\,s per keyframe~\cite{rosinol2021kimera}.
Among systems whose graph feeds a downstream task directly, S\nobreakdash-Graphs+
reports a 74--263\,ms back-end while driving navigation~\cite{bavle2023sgraphs}, which
S\nobreakdash-Graphs 2.0 brings down to an average of 34\,ms~\cite{bavle2025sgraphs2},
and Clio builds a task-driven 3DSG at 0.26--0.31\,s per frame on a laptop carried by a
quadruped~\cite{maggio2024clio}. On the pure SLAM side, the tracking thread of
ORB\nobreakdash-SLAM2 runs in 26--49\,ms per frame~\cite{murartal2017orbslam2} and that
of ORB\nobreakdash-SLAM3 in 22--33\,ms, the latter reporting that it ``is able to run
in real time at 30-40 frames and at 3-6 keyframes per second'' with a mapping thread
that takes 159--267\,ms per keyframe~\cite{campos2021orbslam3}.

Two budgets therefore bound the space these numbers describe, the 25\,ms per-frame
budget of a SLAM front end and the 200\,ms keyframe period in which a 3DSG consumer
actually lives. Every query we measure on \chocolate, \mint, and \saffron falls
between 2.06\,\si{\micro\second} and 126.3\,\si{\micro\second}, which is two to five
orders of magnitude below either figure. The most expensive of them consumes 0.5\% of
a 25\,ms tracking frame and 0.06\% of the 200\,ms keyframe budget, leaving room for
$10^3$ to $10^4$ queries per cycle. It is this margin, rather than the cost of a
single query, that matters in practice: one query fits every budget above, \dsg's
included, and what changes is how many fit. A consumer that probes visibility at one
thousand candidate poses per cycle spends 6.7\,ms on \mint but 350\,ms on \dsg on the
Jetson, the latter exhausting the keyframe budget on its own before any perception
work is accounted
for~\cite{viswanathan2025spadescalablepathplanning, ejaz2025situationallyawarepathplanningexploiting}.

Each \ygg implementation also offers a distinct set of trade-offs a user can choose
from. \chocolate balances low memory consumption against reasonably fast query times,
while \mint and \saffron buy further speed at the price of memory. The choice between
those two concerns the upstream as much as the downstream: \saffron's fine-grained
physical layer allows easier integration with graph generation, since a single point
can be modified without rewriting an array, whereas \mint keeps a memory footprint on
par with \dsg but lacks that precision for operations such as graph update and graph
merge. The cost of that precision is visible in update latency, where a single update
takes 6.91\,\si{\micro\second} on \chocolate, 12.7\,\si{\micro\second} on \mint, and
129\,\si{\micro\second} on \saffron.

\section{Case Studies}
\label{sec:integration}

We integrated \ygg into three published robotic and perception projects. The three
span the perception pipeline rather than one side of it. LP2 is purely downstream,
reading a graph that is built once and never changes. OSG exercises both halves at
once, constructing its graph and querying it inside the same loop that chooses the
robot's next action. S3DSG sits between them, overlaying a second, human-centric
layer onto a graph that already exists and then querying the result. They differ
equally in what \ygg displaces: the conversion step in LP2, the entire store in OSG,
and a hand-rolled family of spatial indexes in S3DSG.

In each project we replaced only the scene-graph interaction and left the
application logic untouched, then ran both configurations so that the \ygg version
could be checked against the original. What that check can consist of differs with
the project, and we report it in each; what we are after is the performance
difference, that is, how much computation a scene graph designed for consumption
removes from a pipeline that was not built around one.

\subsection{LP2: Long-Term Human Trajectory Prediction}
\label{sec:integration:lp2}

LP2~\cite{gorlo2024trajectory} predicts a distribution over a person's position up
to 60 seconds ahead, using an LLM to guess upcoming human-object interactions and a
3D scene graph to ground them in space. It only ever reads its graph, making it the
clean downstream case.

\paragraph{\textbf{Setup}} LP2 ships with two scenes: the office graph used in
Section~\ref{sec:eval}, with 45 trajectories, and a home graph of 861 waypoints, 110
objects, and 31 trajectories. We compare its original backend, a mix of
DSG~\cite{rosinol2020dynamic, hughes2024foundations} and
NetworkX~\cite{hagberg2008networkx}, against an \ygg one. Because LP2 queries an LLM
internally, we recorded every response from
\texttt{gpt-5.6-luna}~\cite{openai2026gpt56} on a first run so that later runs replay
them rather than calling a live model, and each configuration then runs five times per
scene. Correctness holds at all three levels we check: \ygg stores every node, edge,
and relation the original does, the two variants produce byte-identical LLM prompts,
and the predicted trajectory distributions match on every run.

\paragraph{\textbf{Performance}} Overall, \ygg removes 99.6\% of the time LP2 spends
on its scene graph in the office scene and 98.8\% in the home scene
(Table~\ref{tab:lp2}). Two distinct factors account for it.

The first is query runtime. The clearest case is the waypoint-to-room query, which
at over 1.7\,M calls on the office scene is by a wide margin the most frequent in the
pipeline, and which \ygg answers $873\times$ faster there and $636\times$ faster on
the home scene. The reason is structural: \ygg holds that relationship as a nesting
edge and reads it as a single parent pointer, whereas the original must iterate the
entire edge list to find the edge joining a waypoint to its room. The original's cost
therefore scales with the graph while \ygg's does not, and the gap widens as the scene
grows. \ygg's indexing does the same for
spatial queries, answering radius queries $107\times$ (office) and $87\times$ (home)
faster and nearest-waypoint queries $30\times$ and $25\times$ faster.

The second factor is the removal of the second representation the
original must build before it can ask one. DSG cannot answer the queries LP2 needs,
so the pipeline re-materializes its graph as a NetworkX copy and rebuilds a KD-tree
every iteration from the upstream 3DSG for the radius queries. Building those structures costs \SI{2681}{\milli\second} per run
on the office scene and \SI{1355}{\milli\second} on the home scene, against
\SI{35}{\milli\second} and \SI{27}{\milli\second} for \ygg. The saving is worth more
than the milliseconds: conversion is viable only while the graph stays frozen, and on
a deployed robot, where the graph is updated while a downstream task runs, it must be
repaid after every update.

\begin{table}[!t]
\caption{Scene-graph query cost in LP2, pooled over five repetitions per scene.
The five query rows are median per-call latencies in \si{\micro\second}; the two
aggregate rows below them are per-run totals in \si{\milli\second}. The original
builds its NetworkX copy and KD-tree at startup and rebuilds the KD-tree on every
iteration, whereas \ygg builds its indexes once and maintains them incrementally.}
\label{tab:lp2}
\centering
\small
\setlength{\tabcolsep}{2.4pt}
\begin{tabular}{l rr rr}
\toprule
 & \multicolumn{2}{c}{\textbf{office}} & \multicolumn{2}{c}{\textbf{home}} \\
\cmidrule(lr){2-3}\cmidrule(lr){4-5}
query & orig. & \ygg & orig. & \ygg \\
\midrule
room of a waypoint      & 987   & \textbf{1.13} & 725   & \textbf{1.14} \\
nearest waypoint        & 169   & 5.7           & 134   & 5.3 \\
waypoints within 0.5\,m & 8,420 & 79            & 5,981 & 69 \\
shortest-path lookup    & 0.50  & 0.63          & 0.50  & 0.65 \\
grounding distances     & 20    & 23            & 14    & 14 \\
\midrule
build time              & 2,681 & 35            & 1,355 & 27 \\
\textbf{total graph time} & \textbf{7,747} & \textbf{31}
                          & \textbf{1,752} & \textbf{21} \\
\bottomrule
\end{tabular}
\end{table}

\subsection{OSG: Zero-shot LLM-Guided Object-Search Navigation}
\label{sec:integration:osg}

Open Scene Graph (OSG)~\cite{loo2024osg} performs object-goal navigation, with an
LLM reasoning over a scene graph built progressively as the agent explores and
yields the next action. Unlike LP2, OSG builds and queries the graph in the same
iteration, which makes it the case where upstream construction and downstream
consumption meet inside the control loop.

\paragraph{\textbf{Setup}} We compare the original OSG backend against a \vanilla
\ygg backend over 400 episodes, 20 in each of 20 Habitat-simulated HM3D indoor
scenes~\cite{ramakrishnan2021hm3d}, in ground-truth mode where object detection is
provided by the simulator's semantic mesh. As with LP2 we cached the responses of the
LLM that OSG queries, \texttt{gpt-5.6-luna}~\cite{openai2026gpt56}, keyed by prompt
content, so that the two backends stay comparable despite the stochasticity of live
feedback. Because the graph state at step $n$ depends on the action taken at step
$n-1$, correctness here means the full sequence of actions and positions matching step
by step; across all 400 episodes there was zero divergence, with identical mean
success, SPL, and step count.

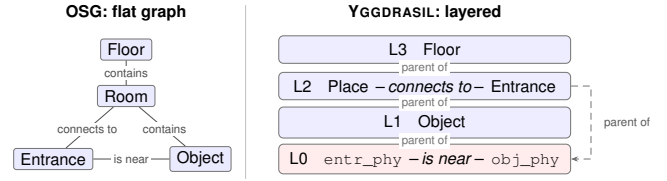
\begin{figure}[t!]
  \centering
  \resizebox{\columnwidth}{!}{
\begin{tikzpicture}[
  x=1cm, y=1cm, font=\sffamily\scriptsize,
  nd/.style   = {draw=black!55, rounded corners=1.2pt, fill=blue!8,
                 inner xsep=3pt, inner ysep=2pt, align=center},
  band/.style = {draw=black!55, rounded corners=2pt, fill=blue!7,
                 minimum height=0.40cm, minimum width=4.4cm, inner xsep=4pt},
  phyb/.style = {draw=black!55, rounded corners=2pt, fill=red!7,
                 minimum height=0.40cm, minimum width=4.4cm, inner xsep=4pt},
  lbl/.style  = {fill=white, inner sep=1pt, font=\sffamily\tiny},
  rel/.style  = {black!70, thin},
  par/.style  = {-{Stealth[length=3pt]}, black!55, thin},
  nest/.style = {-{Stealth[length=3pt]}, black!55, thin, densely dashed},
]
\node[font=\sffamily\scriptsize\bfseries] at (1.50,2.60) {OSG: flat graph};
\node[nd] (f) at (1.50,2.05) {Floor};
\node[nd] (r) at (1.50,1.35) {Room};
\node[nd] (e) at (0.40,0.40) {Entrance};
\node[nd] (o) at (2.62,0.40) {Object};
\draw[rel] (f) -- node[lbl] {contains} (r);
\draw[rel] (r) -- node[lbl,pos=0.55] {connects to} (e);
\draw[rel] (r) -- node[lbl,pos=0.55] {contains} (o);
\draw[rel] (e) -- node[lbl] {is near} (o);
\draw[black!25] (3.30,0.10) -- (3.30,2.72);
\node[font=\sffamily\scriptsize\bfseries] at (6.00,2.60) {\textsc{Yggdrasil}: layered};
\node[band] (l3) at (6.00,2.05) {L3\hspace{5pt} Floor};
\node[band] (l2) at (6.00,1.50)
      {L2\hspace{5pt} Place \,--\,\textit{connects to}\,--\, Entrance};
\node[band] (l1) at (6.00,0.95) {L1\hspace{5pt} Object};
\node[phyb] (l0) at (6.00,0.40)
      {L0\hspace{5pt} \texttt{entr\_phy} \,--\,\textit{is near}\,--\, \texttt{obj\_phy}};
\draw[par] (l3) -- node[lbl] {parent of} (l2);
\draw[par] (l2) -- node[lbl] {parent of} (l1);
\draw[par] (l1) -- node[lbl] {parent of} (l0);
\draw[nest] (l2.east) -- ++(0.30,0) |- (l0.east);
\node[font=\sffamily\tiny, black!65, anchor=west] at (8.58,0.95) {parent of};
\end{tikzpicture}}
  \caption{OSG's flat scene graph (left) and the same scene in \ygg's layers
  (right); the \emph{is near} relation moves down to the shadow nodes in the
  physical layer.}
  \label{fig:osg}
\end{figure}
\paragraph{\textbf{Structure}} OSG keeps its graph in NetworkX, which gives a flat
representation designed with undirected and unweighted edges and whose nodes may carry
several relations at once: an entrance node connects to rooms through a
\emph{connects to} relation and to objects through \emph{is near}. \ygg's structure
is stricter, and mapping OSG onto it required one accommodation.
Fig.~\ref{fig:osg} shows the result, a four-layer design of floor, place and
entrance, object, and physical layers. Because an \ygg edge must join two nodes in
the same layer, the \emph{is near} relation between an entrance and an object cannot
be expressed directly; each entrance and object node therefore carries a shadow node
in the physical layer, created lazily on first use and nested under its owner, and
the relation is held between the two shadow nodes.

\paragraph{\textbf{Performance}} Mean total graph latency falls from
\SI{488.1}{\milli\second} to \SI{229.3}{\milli\second} and memory from 147.9\,MB to
134.5\,MB (Table~\ref{tab:osg-yggdrasil}). The gain concentrates in the read-heavy
operations a typed, layered store is built for: retrieving nodes by type is
$26.7\times$ faster and checking a node's type $14.0\times$. One operation is slower.
Inserting an edge costs $0.49\times$, because \ygg's edges are primitively directed,
so each of OSG's undirected \emph{connects to} and \emph{is near} relations requires
two calls into the engine.
\begin{table}[ht]
    \centering
    \caption{Performance comparison between OSG and Yggdrasil.}
    \label{tab:osg-yggdrasil}
    \small
    \setlength{\tabcolsep}{1.7pt}
    \begin{tabular}{lrrl}
        \toprule
        \textbf{Operation} & \textbf{OSG (ms)} & \textbf{Yggdrasil (ms)} & \textbf{Result} \\
        \midrule
        \textbf{Total graph-update time}
            & \textbf{488.117}
            & \textbf{229.312}
            & \textbf{2.13$\times$} \\
        \quad Serialize graph to prompt
            & 144.310 & 20.622
            & 7.00$\times$ \\
        \quad Get nodes by type
            & 59.451 & 2.223
            & 26.74$\times$ \\
        \quad Check node type
            & 39.903 & 2.860
            & 13.95$\times$ \\
        \quad Add node
            & 83.763 & 54.467
            & 1.54$\times$ \\
        \quad Count instances
            & 158.920 & 145.510
            & 1.09$\times$ \\
        \quad Add Edge
            & 1.770 & 3.630
            & 0.49$\times$ \\
        \midrule
    \textbf{Memory}
        & 147.9 MB & \textbf{134.5 MB} & \textbf{9\% less} \\

    \bottomrule
    \end{tabular}
\end{table}

\subsection{S3DSG: Social 3D Scene Graphs for Human-Aware Reasoning}
\label{sec:Social3DSG}

Social 3D Scene Graphs (S3DSG)~\cite{bartoli2025social} augment a conventional 3DSG
with humans and the activities relating them to objects and to one another.
\hide{
, inferred
by a module the authors call ReaSoN, which resolves even activities whose object is
off-screen by reasoning about head pose and visibility
}
The resulting graph answers
spatial, activity, and functional queries, and drives a socially-aware
motion-planning task in which an identified interaction between two people becomes a
cost field the planner must avoid crossing.

S3DSG overlays a
second, human-centric layer of nodes and activity edges onto an already-populated
3DSG, precisely the kind of layer \ygg is designed to add without disturbing the
ones beneath it (Section~\ref{sec:design}). Its three query categories map onto
\ygg's API directly: spatial queries to \texttt{nearest\_nodes} and
\texttt{nodes\_within\_radius}, activity queries to \texttt{edges\_matching} and
\texttt{nodes\_having}, and occlusion estimator with \ygg's \texttt{field-of-view} query.

\paragraph{\textbf{Setup}} Since the original pipeline already maintains its own spatial indexes, what
this case study isolates is \ygg's layer indexing, measured against a hand-rolled
external equivalent. The scene is a reconstructed apartment of 83 object nodes and 84
relation edges. The original keeps it in NetworkX~\cite{hagberg2008networkx} with a
Open3D~\cite{zhou2018open3d} cloud on every node and one
SciPy~\cite{virtanen2020scipy} \texttt{cKDTree} per object plus one over the floor,
while \ygg expresses the same scene as three position-indexed metric layers of one
node per point, nested beneath a semantic layer of 80 object nodes carrying the
activity edges. Evaluating the cost field issues \num{2115650} nearest-neighbour
queries; the store itself is not under test, since traversing its 84 edges costs
\SI{41}{\micro\second}. Both versions agree on the cost field to within
\num{3.2e-6}.

\paragraph{\textbf{Performance}} \ygg evaluates the cost field in
\SI{4936}{\milli\second} against \SI{6885}{\milli\second} for the original,
a speedup of \num{1.39} that holds to within two percent across runs, and it reuses
the metric layer it already stores rather than rebuilding the renderable cloud on
each evaluation, saving a further \SI{1343}{\milli\second}. The layer decomposition
accounts for the gap, since both rely on kd-trees. Because \ygg's metric data is
partitioned by layer, one index per layer suffices and a query descends a single tree,
whereas the original, whose indexes are keyed by object, must probe one tree per object
and reduce the answers to a minimum, so its cost grows with the number of indexed
objects while \ygg's does not.

The margin is modest, and deliberately so. S3DSG is the case in which the original
already does by hand what \ygg does internally, maintaining a family of kd-tree
indexes over the metric data and keeping them in step with the scene. It is thus
close to the best an external, hand-rolled indexing layer can do, and a large speedup
was never the outcome to expect. What it shows is that even against that best case,
moving the indexes inside the scene graph gains performance rather than costing it,
and does so while the application writes and maintains none of the indexing code and
the indexes stay live as the graph is updated.

\section{Conclusion}
\label{sec:conclusion}

We presented \ygg, a layer-first 3D scene graph built from generic nodes, edges, and
layers, in which the layer hierarchy is a directed acyclic graph rather than a stack
and a typed query interface serves semantic and spatial queries in place, without an
intermediate conversion. We integrated it into three published pipelines spanning
human trajectory prediction, object-goal navigation, and human-aware motion planning,
which between them cover downstream-only use, construction and consumption together
inside a control loop, and augmentation of a graph that already exists.

\ygg is compatible with the common generation frameworks perception pipelines already
rely on, but compatibility is not optimization. Two gaps remain: \chocolate's per-room
schema was designed by hand, and \ygg has no concurrency control, since every
measurement here is single-threaded. Tuning update and merge paths to the access
patterns construction actually produces, inferring schemas automatically, and
supporting concurrent readers and writers are our directions for future work.

\bibliographystyle{IEEEtran}
\bibliography{references}

@inproceedings{armeni2019scenegraph,
  author       = {Iro Armeni and
                  Zhi{-}Yang He and
                  Amir Zamir and
                  JunYoung Gwak and
                  Jitendra Malik and
                  Martin Fischer and
                  Silvio Savarese},
  title        = {3D Scene Graph: {A} Structure for Unified Semantics, 3D Space, and
                  Camera},
  booktitle    = {2019 {IEEE/CVF} International Conference on Computer Vision, {ICCV}
                  2019, Seoul, Korea (South), October 27 - November 2, 2019},
  pages        = {5663--5672},
  publisher    = {{IEEE}},
  year         = {2019},
  url          = {https://doi.org/10.1109/ICCV.2019.00576},
  doi          = {10.1109/ICCV.2019.00576}
}

@inproceedings{rosinol2020dynamic,
  author    = {Antoni Rosinol and Arjun Gupta and Marcus Abate and Jingnan Shi and Luca Carlone},
  title     = {{3D Dynamic Scene Graphs: Actionable Spatial Perception with Places, Objects, and Humans}},
  booktitle = {Proceedings of Robotics: Science and Systems},
  year      = {2020},
  address   = {Corvalis, Oregon, USA},
  month     = {July},
  doi       = {10.15607/RSS.2020.XVI.079}
}

@article{rosinol2021kimera,
  author       = {Antoni Rosinol and
                  Andrew Violette and
                  Marcus Abate and
                  Nathan Hughes and
                  Yun Chang and
                  Jingnan Shi and
                  Arjun Gupta and
                  Luca Carlone},
  title        = {Kimera: From {SLAM} to spatial perception with 3D dynamic scene graphs},
  journal      = {Int. J. Robotics Res.},
  volume       = {40},
  number       = {12-14},
  pages        = {1510--1546},
  year         = {2021},
  url          = {https://doi.org/10.1177/02783649211056674},
  doi          = {10.1177/02783649211056674}
}

@inproceedings{hughes2022hydra,
  author    = {Nathan Hughes and Yun Chang and Luca Carlone},
  title     = {{Hydra: A Real-time Spatial Perception System for 3D Scene Graph Construction and Optimization}},
  booktitle = {Proceedings of Robotics: Science and Systems},
  year      = {2022},
  address   = {New York City, NY, USA},
  month     = {June},
  doi       = {10.15607/RSS.2022.XVIII.050}
}

@inproceedings{rana2023sayplan,
  title     = {SayPlan: Grounding Large Language Models using 3D Scene Graphs for Scalable Robot Task Planning},
  author    = {Rana, Krishan and Haviland, Jesse and Garg, Sourav and Abou-Chakra, Jad and Reid, Ian and Suenderhauf, Niko},
  booktitle = {Proceedings of The 7th Conference on Robot Learning},
  pages     = {23--72},
  year      = {2023},
  editor    = {Tan, Jie and Toussaint, Marc and Darvish, Kourosh},
  volume    = {229},
  series    = {Proceedings of Machine Learning Research},
  month     = {06--09 Nov},
  publisher = {PMLR},
  url       = {https://proceedings.mlr.press/v229/rana23a.html}
}

@inproceedings{gu2024conceptgraphs,
  author       = {Qiao Gu and
                  Ali Kuwajerwala and
                  Sacha Morin and
                  Krishna Murthy Jatavallabhula and
                  Bipasha Sen and
                  Aditya Agarwal and
                  Corban Rivera and
                  William Paul and
                  Kirsty Ellis and
                  Rama Chellappa and
                  Chuang Gan and
                  Celso Miguel de Melo and
                  Joshua B. Tenenbaum and
                  Antonio Torralba and
                  Florian Shkurti and
                  Liam Paull},
  title        = {ConceptGraphs: Open-Vocabulary 3D Scene Graphs for Perception and
                  Planning},
  booktitle    = {{IEEE} International Conference on Robotics and Automation, {ICRA}
                  2024, Yokohama, Japan, May 13-17, 2024},
  pages        = {5021--5028},
  publisher    = {{IEEE}},
  year         = {2024},
  url          = {https://doi.org/10.1109/ICRA57147.2024.10610243},
  doi          = {10.1109/ICRA57147.2024.10610243}
}

@article{maggio2024clio,
  author       = {Dominic Maggio and
                  Yun Chang and
                  Nathan Hughes and
                  Matthew Trang and
                  J. Daniel Griffith and
                  Carlyn Dougherty and
                  Eric Cristofalo and
                  Lukas Schmid and
                  Luca Carlone},
  title        = {Clio: Real-Time Task-Driven Open-Set 3D Scene Graphs},
  journal      = {{IEEE} Robotics Autom. Lett.},
  volume       = {9},
  number       = {10},
  pages        = {8921--8928},
  year         = {2024},
  url          = {https://doi.org/10.1109/LRA.2024.3451395},
  doi          = {10.1109/LRA.2024.3451395}
}

@inproceedings{saxena2024grapheqa,
  title     = {GraphEQA: Using 3D Semantic Scene Graphs for Real-time Embodied Question Answering},
  author    = {Saxena, Saumya and Buchanan, Blake and Paxton, Chris and Liu, Peiqi and Chen, Bingqing and Vaskevicius, Narunas and Palmieri, Luigi and Francis, Jonathan and Kroemer, Oliver},
  booktitle = {Proceedings of The 9th Conference on Robot Learning},
  pages     = {2714--2742},
  year      = {2025},
  editor    = {Lim, Joseph and Song, Shuran and Park, Hae-Won},
  volume    = {305},
  series    = {Proceedings of Machine Learning Research},
  month     = {27--30 Sep},
  publisher = {PMLR},
  url       = {https://proceedings.mlr.press/v305/saxena25a.html}
}

@article{hughes2024foundations,
  author       = {Nathan Hughes and
                  Yun Chang and
                  Siyi Hu and
                  Rajat Talak and
                  Rumaisa Abdulhai and
                  Jared Strader and
                  Luca Carlone},
  title        = {Foundations of spatial perception for robotics: Hierarchical representations
                  and real-time systems},
  journal      = {Int. J. Robotics Res.},
  volume       = {43},
  number       = {10},
  pages        = {1457--1505},
  year         = {2024},
  url          = {https://doi.org/10.1177/02783649241229725},
  doi          = {10.1177/02783649241229725}
}

@inproceedings{agia2022taskography,
  title = {TASKOGRAPHY: Evaluating robot task planning over large 3D scene graphs},
  author = {Agia, Christopher and Jatavallabhula, Krishna Murthy and Khodeir, Mohamed and Miksik, Ondrej and Vineet, Vibhav and Mukadam, Mustafa and Paull, Liam and Shkurti, Florian},
  booktitle = {Conf. on Robot Learning (CoRL)},
  year = {2021}
}

@article{bae2023survey3dscenegraphs,
  title = {A Survey on 3D Scene Graphs: Definition, Generation and Application},
  author = {Bae, Jaewon and Shin, Dongmin and Ko, Kangbeen and Lee, Juchan and Kim, Ue-Hwan},
  journal = {Robot Intelligence Technology and Applications},
  year = {2022}
}

@article{bavle2022situational,
  title = {Situational Graphs for Robot Navigation in Structured Indoor Environments},
  author = {Bavle, Hriday and Sanchez-Lopez, Jose Luis and Shaheer, Muhammad and Civera, Javier and Voos, Holger},
  journal = {IEEE Robotics and Automation Letters},
  year = {2022}
}

@article{bavle2023sgraphs,
  title = {S-Graphs+: Real-time Localization and Mapping leveraging Hierarchical Representations},
  author = {Bavle, Hriday and Sanchez-Lopez, Jose Luis and Shaheer, Muhammad and Civera, Javier and Voos, Holger},
  journal = {IEEE Robotics and Automation Letters},
  year = {2023}
}

@article{bavle2025sgraphs2,
  title = {S-Graphs 2.0 – A Hierarchical-Semantic Optimization and Loop Closure for SLAM},
  author = {Bavle, Hriday and Sanchez-Lopez, Jose Luis and Shaheer, Muhammad and Civera, Javier and Voos, Holger},
  journal = {IEEE Robotics and Automation Letters},
  year = {2025}
}

@article{deng2024opengraph,
  title = {OpenGraph: Open-Vocabulary Hierarchical 3D Graph Representation in Large-Scale Outdoor Environments},
  author = {Deng, Yinan and Wang, Jiahui and Zhao, Jingyu and Tian, Xinyu and Chen, Guangyan and Yang, Yi and Yue, Yufeng},
  journal = {IEEE Robotics and Automation Letters},
  year = {2024}
}

@article{gorlo2024trajectory,
  title = {Long-Term Human Trajectory Prediction using 3D Dynamic Scene Graphs},
  author = {Gorlo, Nicolas and Schmid, Lukas and Carlone, Luca},
  journal = {IEEE Robotics and Automation Letters},
  year = {2024}
}

@article{honerkamp2024language,
  title = {Language-Grounded Dynamic Scene Graphs for Interactive Object Search with Mobile Manipulation},
  author = {Honerkamp, Daniel and Büchner, Martin and Despinoy, Fabien and Welschehold, Tim and Valada, Abhinav},
  journal = {IEEE Robotics and Automation Letters},
  year = {2024}
}

@article{kim2019sparse3d,
  title = {3-D Scene Graph: A Sparse and Semantic Representation of Physical Environments for Intelligent Agents},
  author = {Kim, Ue-Hwan and Park, Jin-Man and Song, Taek-Jin and Kim, Jong-Hwan},
  journal = {IEEE Trans. on Cybernetics},
  year = {2019}
}

@inproceedings{koch2024open3dsg,
  title = {Open3DSG: Open-Vocabulary 3D Scene Graphs from Point Clouds with Queryable Objects and Open-Set Relationships},
  author = {Koch, Sebastian and Vaskevicius, Narunas and Colosi, Mirco and Hermosilla, Pedro and Ropinski, Timo},
  booktitle = {IEEE Conf. on Computer Vision and Pattern Recognition (CVPR)},
  year = {2024}
}

@inproceedings{linok2025beyondbarequeries,
  title = {Beyond Bare Queries: Open-Vocabulary Object Grounding with 3D Scene Graph},
  author = {Linok, Sergey and Zemskova, Tatiana and Ladanova, Svetlana and Titkov, Roman and Yudin, Dmitry and Monastyrny, Maxim and Valenkov, Aleksei},
  booktitle = {IEEE Intl. Conf. on Robotics and Automation (ICRA)},
  year = {2025}
}

@inproceedings{liu2025delta,
  title = {DELTA: Decomposed Efficient Long-Term Robot Task Planning using Large Language Models},
  author = {Liu, Yuchen and Palmieri, Luigi and Koch, Sebastian and Georgievski, Ilche and Aiello, Marco},
  booktitle = {IEEE Intl. Conf. on Robotics and Automation (ICRA)},
  year = {2025}
}

@article{loo2024osg,
  author  = {Joel Loo and Zhanxin Wu and David Hsu},
  title   = {Open Scene Graphs for Open-World Object-Goal Navigation},
  journal = {The International Journal of Robotics Research},
  year    = {2025}
}

@inproceedings{ramakrishnan2021hm3d,
  title={Habitat-Matterport 3D Dataset ({HM}3D): 1000 Large-scale 3D Environments for Embodied {AI}},
  author={Santhosh Kumar Ramakrishnan and Aaron Gokaslan and Erik Wijmans and Oleksandr Maksymets and Alexander Clegg and John M Turner and Eric Undersander and Wojciech Galuba and Andrew Westbury and Angel X Chang and Manolis Savva and Yili Zhao and Dhruv Batra},
  booktitle={Thirty-fifth Conference on Neural Information Processing Systems Datasets and Benchmarks Track},
  year={2021},
  url={https://arxiv.org/abs/2109.08238}
}

@misc{mahdi2026scout,
  title = {Relational Semantic Reasoning on 3D Scene Graphs for Open World Interactive Object Search},
  author = {Mahdi, Imen and Cassinelli, Matteo and Despinoy, Fabien and Welschehold, Tim and Valada, Abhinav},
  archivePrefix = {arXiv preprint},
  year = {2026}
}

@misc{ra2025structuredinterface,
  title = {Structured Interfaces for Automated Reasoning with 3D Scene Graphs},
  author = {Ray, Aaron and Arkin, Jacob and Biggie, Harel and Fan, Chuchu and Carlone, Luca and Roy, Nicholas},
  archivePrefix = {arXiv preprint},
  year = {2025}
}

@misc{rotondi2026survey,
  title = {3D Scene Graphs: Open Challenges and Future Directions},
  author = {Rotondi, Dennis and Argenziano, Francesco and Koch, Sebastian and Hughes, Nathan and Büchner, Martin and Wald, Johanna and Rosenberger Schmid, Lukas and Nardi, Daniele and Valada, Abhinav and Paull, Liam and Tombari, Federico and Carlone, Luca and Arras, Kai O.},
  year = {2026},
  eprint = {2606.19383},
  archivePrefix = {arXiv}
}

@inproceedings{wald2020learning3d,
  title = {Learning 3D Semantic Scene Graphs from 3D Indoor Reconstructions},
  author = {Wald, Johanna and Dhamo, Helisa and Navab, Nassir and Tombari, Federico},
  booktitle = {IEEE Conf. on Computer Vision and Pattern Recognition (CVPR)},
  year = {2020}
}

@article{werby2024hierarchical,
  title = {Hierarchical Open-Vocabulary 3D Scene Graphs for Language-Grounded Robot Navigation},
  author = {Werby, Abdelrhman and Huang, Chenguang and Büchner, Martin and Valada, Abhinav and Burgard, Wolfram},
  journal = {Robotics: Science and Systems (RSS)},
  year = {2024}
}

@inproceedings{werby2025keysg,
  title = {KeySG: Hierarchical Keyframe-Based 3D Scene Graphs},
  author = {Werby, Abdelrhman and Rotondi, Dennis and Scaparro, Fabio and Arras, Kai O.},
  booktitle = {IEEE Intl. Conf. on Robotics and Automation (ICRA)},
  year = {2026}
}

@inproceedings{wu2021scenegraphfusion,
  title = {SceneGraphFusion: Incremental 3D Scene Graph Prediction from RGB-D Sequences},
  author = {Wu, Shun-Cheng and Wald, Johanna and Tateno, Keisuke and Navab, Nassir and Tombari, Federico},
  booktitle = {IEEE Conf. on Computer Vision and Pattern Recognition (CVPR)},
  year = {2021}
}

@article{yin2024sg,
  title = {SG-Nav: Online 3D Scene Graph Prompting for LLM-based Zero-shot Object Navigation},
  author = {Yin, Hang and Xu, Xiuwei and Wu, Zhenyu and Zhou, Jie and Lu, Jiwen},
  journal = {Advances in Neural Information Processing Systems (NeurIPS)},
  year = {2024}
}

@misc{zhou2025fsrvln,
  title = {FSR-VLN: Fast and Slow Reasoning for Vision-Language Navigation with Hierarchical Multi-modal Scene Graph},
  author = {Zhou, Xiaolin and Xiao, Tingyang and Liu, Liu and Wang, Yucheng and Chen, Maiyue and Meng, Xinrui and Wang, Xinjie and Feng, Wei and Sui, Wei and Su, Zhizhong},
  archivePrefix = {arXiv preprint},
  year = {2025}
}

@article{bentley1975kdtree,
  author  = {Jon Louis Bentley},
  title   = {Multidimensional Binary Search Trees Used for Associative Searching},
  journal = {Communications of the {ACM}},
  volume  = {18},
  number  = {9},
  pages   = {509--517},
  year    = {1975},
  doi     = {10.1145/361002.361007}
}

@article{friedman1977kdtree,
  author  = {Jerome H. Friedman and Jon Louis Bentley and Raphael Ari Finkel},
  title   = {An Algorithm for Finding Best Matches in Logarithmic Expected Time},
  journal = {{ACM} Transactions on Mathematical Software},
  volume  = {3},
  number  = {3},
  pages   = {209--226},
  year    = {1977},
  doi     = {10.1145/355744.355745}
}

@misc{sparkdsg,
  author       = {{MIT SPARK Lab}},
  title        = {{Spark-DSG}: A Library for Dynamic Scene Graphs},
  year         = {2022},
  howpublished = {\url{https://github.com/MIT-SPARK/Spark-DSG}},
  note         = {Open-source software; distributes the \texttt{uHumans2}
                  office example scene graph. Accessed 2026-09-09}
}

@inproceedings{zhou2017ade20k,
  author    = {Bolei Zhou and Hang Zhao and Xavier Puig and Sanja Fidler and
               Adela Barriuso and Antonio Torralba},
  title     = {Scene Parsing through {ADE20K} Dataset},
  booktitle = {{IEEE} Conf.\ on Computer Vision and Pattern Recognition (CVPR)},
  year      = {2017},
  doi       = {10.1109/CVPR.2017.544}
}

@article{campos2021orbslam3,
  title   = {{ORB-SLAM3}: An Accurate Open-Source Library for Visual,
             Visual-Inertial, and Multimap {SLAM}},
  author  = {Campos, Carlos and Elvira, Richard and G{\'o}mez Rodr{\'i}guez, Juan J.
             and Montiel, Jos{\'e} M. M. and Tard{\'o}s, Juan D.},
  journal = {IEEE Transactions on Robotics},
  volume  = {37},
  number  = {6},
  pages   = {1874--1890},
  year    = {2021},
  doi     = {10.1109/TRO.2021.3075644}
}

@article{murartal2017orbslam2,
  title   = {{ORB-SLAM2}: An Open-Source {SLAM} System for Monocular, Stereo,
             and {RGB-D} Cameras},
  author  = {Mur-Artal, Ra{\'u}l and Tard{\'o}s, Juan D.},
  journal = {IEEE Transactions on Robotics},
  volume  = {33},
  number  = {5},
  pages   = {1255--1262},
  year    = {2017},
  doi     = {10.1109/TRO.2017.2705103}
}

@article{cadena2016past,
  author  = {Cesar Cadena and Luca Carlone and Henry Carrillo and Yasir Latif and
             Davide Scaramuzza and Jos{\'e} Neira and Ian Reid and John J. Leonard},
  title   = {Past, Present, and Future of Simultaneous Localization and Mapping:
             Toward the Robust-Perception Age},
  journal = {IEEE Transactions on Robotics},
  volume  = {32},
  number  = {6},
  pages   = {1309--1332},
  year    = {2016},
  doi     = {10.1109/TRO.2016.2624754}
}

@article{garg2020semantics,
  author  = {Sourav Garg and Niko S{\"u}nderhauf and Feras Dayoub and
             Douglas Morrison and Akansel Cosgun and Gustavo Carneiro and
             Qi Wu and Tat-Jun Chin and Ian Reid and Stephen Gould and
             Peter Corke and Michael Milford},
  title   = {Semantics for Robotic Mapping, Perception and Interaction: A Survey},
  journal = {Foundations and Trends in Robotics},
  volume  = {8},
  number  = {1--2},
  pages   = {1--224},
  year    = {2020},
  doi     = {10.1561/2300000059}
}

@inproceedings{hagberg2008networkx,
  author    = {Aric A. Hagberg and Daniel A. Schult and Pieter J. Swart},
  title     = {Exploring Network Structure, Dynamics, and Function using {NetworkX}},
  booktitle = {Proceedings of the 7th Python in Science Conference (SciPy)},
  editor    = {Ga{\"e}l Varoquaux and Travis Vaught and Jarrod Millman},
  address   = {Pasadena, CA, USA},
  pages     = {11--15},
  year      = {2008}
}

@misc{ejaz2025situationallyawarepathplanningexploiting,
  title = {Situationally-aware Path Planning Exploiting 3D Scene Graphs},
  author = {Ejaz, Saad and Giberna, Marco and Shaheer, Muhammad and Millan-Romera, Jose Andres and Tourani, Ali and Kremer, Paul and Voos, Holger and Sanchez-Lopez, Jose Luis},
  archivePrefix = {arXiv preprint},
  year = {2025}
}

@inproceedings{viswanathan2025spadescalablepathplanning,
  title = {SPADE: Towards Scalable Path Planning Architecture on Actionable Multi-Domain 3D ScenE Graphs},
  author = {Viswanathan, Vignesh Kottayam and Patel, Akash and Saucedo, Mario Alberto Valdes and Satpute, Sumeet and Kanellakis, Christoforos and Nikolakopoulos, George},
  booktitle = {IEEE/RSJ Intl. Conf. on Intelligent Robots and Systems (IROS)},
  year = {2025}
}

@misc{openai2026gpt56,
  author       = {{OpenAI}},
  title        = {{GPT-5.6}: Frontier Intelligence that Scales with Your Ambition},
  year         = {2026},
  howpublished = {\url{https://openai.com/index/gpt-5-6/}},
  note         = {Model \texttt{gpt-5.6-luna}; accessed 2026-08-28}
}

@article{chang2023dlite,
  title   = {D-Lite: Navigation-Oriented Compression of 3D Scene Graphs for Multi-Robot Collaboration},
  author  = {Chang, Yun and Ebadi, Kamak and Denniston, Christopher E. and Ginting, Muhammad Fadhil and Rosinol, Antoni and Reinke, Andrzej and Palieri, Matteo and Shi, Jingnan and Chatterjee, Arghya and Morrell, Benjamin and Agha-mohammadi, Ali-akbar and Carlone, Luca},
  journal = {IEEE Robotics and Automation Letters},
  year    = {2023}
}

@article{bartoli2025social,
  title   = {Social 3D Scene Graphs: Modeling Human Actions and Relations for Interactive Service Robots},
  author  = {Bartoli, Ermanno and Rotondi, Dennis and He, Buwei and Jensfelt, Patric and Arras, Kai O. and Leite, Iolanda},
  journal = {arXiv preprint arXiv:2509.24966},
  year    = {2025},
  note    = {Accepted at IROS 2026}
}

@misc{zhou2018open3d,
      title={Open3D: A Modern Library for 3D Data Processing},
      author={Qian-Yi Zhou and Jaesik Park and Vladlen Koltun},
      year={2018},
      eprint={1801.09847},
      archivePrefix={arXiv},
      primaryClass={cs.CV},
      url={https://arxiv.org/abs/1801.09847},
}

@article{virtanen2020scipy,
  author  = {Virtanen, Pauli and Gommers, Ralf and Oliphant, Travis E. and
             Haberland, Matt and Reddy, Tyler and Cournapeau, David and
             Burovski, Evgeni and Peterson, Pearu and Weckesser, Warren and
             Bright, Jonathan and van der Walt, St{\'e}fan J. and Brett, Matthew and
             Wilson, Joshua and Millman, K. Jarrod and Mayorov, Nikolay and
             Nelson, Andrew R. J. and Jones, Eric and Kern, Robert and
             Larson, Eric and Carey, C J and Polat, {\.I}lhan and Feng, Yu and
             Moore, Eric W. and VanderPlas, Jake and Laxalde, Denis and
             Perktold, Josef and Cimrman, Robert and Henriksen, Ian and
             Quintero, E. A. and Harris, Charles R. and Archibald, Anne M. and
             Ribeiro, Ant{\^o}nio H. and Pedregosa, Fabian and
             van Mulbregt, Paul and {SciPy 1.0 Contributors}},
  title   = {{SciPy} 1.0: Fundamental Algorithms for Scientific Computing in {P}ython},
  journal = {Nature Methods},
  volume  = {17},
  number  = {3},
  pages   = {261--272},
  year    = {2020},
  doi     = {10.1038/s41592-019-0686-2},
  url     = {https://doi.org/10.1038/s41592-019-0686-2},
}

@article{bartoli2026gesto,
  title={GESTO: Human-Centric Spatio-Temporal Memory for Reasoning in Dynamic Scenes},
  author={Bartoli, Ermanno and He, Buwei and Rotondi, Dennis and Koch, Sebastian and Tombari, Federico and Arras, Kai O and Jensfelt, Patric and Cai, Yixi and Leite, Iolanda},
  journal={arXiv preprint arXiv:2608.10886},
  year={2026}
}

@IEEEtranBSTCTL{IEEEexample:BSTcontrol,
  CTLuse_forced_etal      = "yes",
  CTLmax_names_forced_etal= "3",
  CTLnames_show_etal      = "1"
}

@misc{yggdrasil_artifact,
  title        = {{Yggdrasil}: implementation, benchmark, and integration artifacts},
  howpublished = {Figshare},
  year         = {2026},
  note         = {\url{https://github.com/brdsdsu/yggdrasil}}
}

\end{document}